\documentclass[pmlr]{jmlr}

\RequirePackage{graphicx}
 \usepackage{booktabs}
\usepackage{longtable}
\usepackage{siunitx}

\usepackage[switch]{lineno}
\usepackage{float}
\usepackage{booktabs}
\usepackage{multirow}
\makeatletter
\def\set@curr@file#1{\def\@curr@file{#1}} 
\makeatother
\usepackage[load-configurations=version-1]{siunitx} 

\theorembodyfont{\upshape}
\theoremheaderfont{\scshape}
\theorempostheader{:}
\theoremsep{\newline}

\jmlrvolume{[VOLUME \# TBD]}
\jmlryear{2024}
\jmlrworkshop{Machine Learning for Healthcare}

\title[Representation Learning of Lab Values via Masked AutoEncoders]{Representation Learning of Lab Values via Masked AutoEncoders}

\author{%
\Name{David Restrepo}\footnotemark[1] \\
\addr Massachusetts Institute of Technology (MIT), USA; \addr Université Paris-Saclay, France
\AND
\Name{Chenwei Wu}\footnotemark[1]\\
\addr University of Michigan, USA
\AND
\Name{Yueran Jia}\\
\addr Northeastern University, USA
\AND
\Name{Jaden K. Sun}\\
\addr Massachusetts Institute of Technology (MIT), USA
\AND
\Name{Jack Gallifant}\\
\addr Harvard Medical School, USA; \addr Brigham and Women’s Hospital/Dana-Farber Cancer Institute, USA
\AND
\Name{Catherine G. Bielick}\\
\addr Massachusetts Institute of Technology (MIT), USA; \addr Harvard Medical School, USA; \addr Beth Israel Deaconess Medical Center, USA
\AND
\Name{Yugang Jia}\\
\addr Massachusetts Institute of Technology (MIT), USA
\AND
\Name{Leo A. Celi} \Email{leoanthonyceli@yahoo.com}\\
\addr Massachusetts Institute of Technology (MIT), USA; \addr Harvard Medical School, USA; \addr Beth Israel Deaconess Medical Center, USA
}

\footnotetext[1]{These authors contributed equally to this work.}

\begin{document}

\maketitle

\vspace{-1.5em}

\begin{abstract}
Accurate imputation of missing laboratory values in electronic health records (EHRs) is critical to enable robust clinical predictions and reduce biases in AI systems in healthcare. Existing methods, such as XGBoost, softimpute, GAIN, Expectation Maximization (EM), and MICE, struggle to model the complex temporal and contextual dependencies in EHR data, particularly in underrepresented groups. In this work, we propose Lab-MAE, a novel transformer-based masked autoencoder framework that leverages self-supervised learning for the imputation of continuous sequential lab values. Lab-MAE introduces a structured encoding scheme that jointly models laboratory test values and their corresponding timestamps, enabling explicit capturing temporal dependencies. Empirical evaluation on the MIMIC-IV dataset demonstrates that Lab-MAE significantly outperforms state-of-the-art baselines such as XGBoost, softimpute, GAIN, EM, and MICE across multiple metrics, including root mean square error (RMSE), R-squared (R2), and Wasserstein distance (WD). Notably, Lab-MAE achieves equitable performance across demographic groups of patients, advancing fairness in clinical predictions. We further investigate the role of follow-up laboratory values as potential shortcut features, revealing Lab-MAE's robustness in scenarios where such data is unavailable. The findings suggest that our transformer-based architecture, adapted to the characteristics of EHR data, offers a foundation model for more accurate and fair clinical imputation. In addition, we measure and compare the carbon footprint of Lab-MAE with the a XGBoost model, highlighting its environmental requirements.

\end{abstract}

\newpage

\paragraph*{Data and Code Availability}
The code to reproduce our experiments is available in anonymous \href{https://anonymous.4open.science/r/Lab-MAE-CHIL2025/README.md}{github}.
The data set used is also publicly available\footnote{This paper uses the MIMIC-IV dataset
\citep{johnson2023mimic}, which is available in the PhysioNet repository
\citep{johnson2020mimic}.}



\section{Introduction}
\label{sec:intro}

Laboratory values play a pivotal role in real-time clinical care by demonstrating a patient's baseline physiology, generating a differential diagnosis for acute or chronic illnesses, and guiding prognosis. The effectiveness of machine learning (ML) models in leveraging laboratory data from electronic health records (EHRs) is often hampered by the prevalence of missing values \cite{luo2022evaluating, austin2021missing}, which can severely affect model performance and introduce harmful bias in clinical implementation \cite{riley2024evaluation}. In addition to technical challenges, the social patterning inherent in the data generation, often drives missing data in clinical datasets. Factors such as socioeconomic status, access to healthcare, and systemic biases can significantly influence the availability of laboratory results, affecting underrepresented groups and introducing skew into clinical datasets \cite{teotia2024variation}. Optimal handling of missing data in this context is a critical challenge, as it directly affects the reliability of clinical models in healthcare settings. 

Conventional imputation techniques, such as mean and standard deviation-based substitutions, are not well-suited to clinical tasks due to need of highly contextualized and individualized interpretation over complex data interactions. Other common imputation methods such as Softimpute \cite{softimpute, softimpute_lab}, Expectation Maximization (EM) \cite{li2021imputation}, and Multivariate Imputation by Chained-Equations (MICE) \cite{li2021imputation}, fail to capture the intricate temporal and inter-variable dependencies present in high-dimensional physiological data \cite{li2021imputation}. For example, the clinical importance of high Creatinine values in the hospitalized setting depends heavily on the patient's own baseline values and the presence of any of the potential causes of acute kidney injury, including sepsis, hemorrhage, iatrogenic causes, urinary tract obstruction, and more. Some advanced methods to training multiple tabular models such as XGBoost \cite{chen2016xgboost} or even deep learning methods like GAN based methods such as GAIN \cite{gain, li2021imputation} for individual lab values, have been used to capture these data relations in lab values \cite{zhang2020predicting, chen2016xgboost}. However, even these models often struggle to fully leverage the available information, leading to suboptimal solutions that may overlook valuable contextual details \cite{waljee2013comparison, luo2016using}.

Recent advances in self-supervised learning have opened new avenues for handling missing data by learning robust representations from the available data itself. Previous work on data imputation in EHR datasets has predominantly employed Variational Autoencoders (VAE) \cite{kingma2013auto} due to their ability to model latent distributions and generate plausible imputations \cite{zamanzadeh2021autopopulus}. However, recent studies have shown that Masked Autoencoders (MAEs) offer notable improvements over VAEs, particularly in their ability to reconstruct high-dimensional data with fewer assumptions on the latent space and greater capacity to learn complex feature dependencies directly from the data \cite{he2022masked, bao2021beit}. These advances highlight the potential of MAEs to deliver more accurate and context-aware imputations for clinical data, motivating the development of our proposed framework.

In addition, transformer-based models have shown promise in fields such as natural language processing \cite{devlin2018bert, vaswani2017attention, renc2024zero} and computer vision \cite{dosovitskiy2020image, parvaiz2023vision} due to their ability to model complex patterns and relationships within the data. Despite all the performance and results shown by transformer models in fields such as computer vision and natural language processing, the creation of foundation models and the training of transformer models for tabular data is a field that still needs to be further explored \cite{van2024tabular}. Some recent previous works have shown that deep learning models can improve classical models such as XGBoost \cite{chen2016xgboost} on tabular data tasks \cite{kadra2021well}. Additionally, attention-based methods have demonstrated significant promise in tabular data imputation \cite{lee2023self, wu2020attention}. Kowsar et al. \cite{kowsar2024attention} proposed an attention-based missing value imputation framework that leverages self-attention and between-sample attention mechanisms to reconstruct missing data. Their method surpasses classical machine learning approaches, such as decision-tree-based imputation, and achieves superior performance on several EHR datasets. The potential of self-supervised pretraining for clinical data is immense, especially when combined with the transformer architecture, and techniques such as masked autoencoding, which aim to learn from the inherent structure of the data rather than relying on predefined labels \cite{he2022masked, krishnan2022self}.

Building on this concept, LABRADOR, a novel continuous Transformer model, was designed to model laboratory data by using masked language modeling (MLM) techniques \cite{bellamy2023labrador}. Despite LABRADOR's innovative architecture and its success in capturing continuous lab data patterns, it still faced challenges in consistently outperforming traditional tree-based methods like XGBoost across various downstream tasks. This is in line with a broader body of evidence indicating that deep learning methods often underperform compared to tree-based techniques on tabular data due to a lack of appropriate inductive biases \cite{grinsztajn2022tree}.

Our proposed solution builds on the LABRADOR and ReMasker frameworks \cite{du2023remasker}, introducing a novel masked autoencoder architecture tailored to the unique characteristics of laboratory data. By extending the principles of masked modeling to impute missing values and leveraging temporal information explicitly, our approach addresses the limitations observed in prior methods. Unlike conventional models, our architecture takes timestamps into account, which is critical for clinical data where the sequence of events can significantly influence outcomes.

The main contributions of this paper are as follows:
\begin{enumerate}
    \item We develop and validate a masked autoencoder transformer model specifically designed for imputing missing lab values in EHR data, incorporating temporal and contextual information to enhance imputation accuracy.
    \item We demonstrate that our model not only outperforms state-of-the-art imputation methods, such as XGBoost, softimpute, GAIN, EM, and MICE, in the context of clinical data imputation but also mitigates biases by providing consistent performance across different patient demographics.
    \item We introduce a self-supervised pre-training strategy that effectively learns high-dimensional representations of lab data, paving the way for more robust downstream predictions even in the absence of complete data.
\end{enumerate}

Our approach combines the strengths of masked autoencoding with the Transformer architecture to create a powerful tool for handling missing lab values in clinical datasets. This advancement has the potential to significantly improve the quality and completeness of EHR data, ultimately enhancing clinical decision-making and patient outcomes.
\subsection*{Generalizable Insights about Machine Learning in the Context of Healthcare}

Our work illustrates that a single, foundation-style Transformer model can simultaneously tackle multiple challenges—missing data, bias, and efficiency—within complex clinical datasets. In particular:

\begin{itemize} \item \textbf{Unified Architecture for Tabular Data:} Moving from separate, feature-specific models (e.g., one XGBoost per lab) to a single masked autoencoder not only streamlines deployment but also consistently improves imputation performance. This shift underscores the value of foundation-model thinking for healthcare’s heterogeneous, high-dimensional data. \item \textbf{Fairness Isn’t Necessarily a Trade-Off:} Our findings suggest that high accuracy and equitable performance can be pursued together. By explicitly modeling temporal and contextual factors, we observed stable gains across diverse subgroups, highlighting that fairness can be integrated into design rather than added post hoc. \item \textbf{Efficiency and Carbon Footprint Matter:} As ML models grow more complex, measuring and minimizing emissions becomes crucial—especially in time- and resource-constrained clinical settings. Our approach shows that a carefully architected, single-model strategy can be more environmentally sustainable than multiple, separate predictors. \end{itemize}

\section{Methods}

\subsection{Datasets}

The dataset used in this study is derived from the MIMIC-IV database \cite{johnson2020mimic, johnson2023mimic}, which contains de-identified health records of patients admitted to critical care units at the Beth Israel Deaconess Medical Center between 2008 and 2019. Our focus was on the top 100 most common lab values, selected based on their occurrence in patient records. The cohort comprises data of 1,417,738 stays for training and 100,000 stays for evaluation. These cohorts were extracted and processed using Google BigQuery.

\subsection{Data Processing}

We utilized SQL queries through Google BigQuery to extract lab event data from the labevents table of MIMIC-IV \cite{johnson2020mimic, johnson2023mimic}. The data includes unique hospital admission IDs (hadm\_id), patient race information for a further fairness and bias evaluation, lab test item IDs (itemid), lab test timestamps (charttime), and corresponding numerical lab values (valuenum). The dataset was preprocessed to remove invalid lab values (e.g., negative values) and filtered to include only valid, positive measurements.

For each patient admission, the earliest recorded timestamp for each lab test was used as a reference point. Additional columns were computed to represent the difference between each lab test's timestamp and this reference. The numerical lab values were then normalized using quantile normalization to limit extreme outliers.

We also calculated follow-up values for each lab test (denoted as npval\_last\_{id}) by tracking subsequent tests performed within the same admission. For each lab test and follow-up, a corresponding time difference column was generated (denoted as nptime\_{id}), representing the time elapsed since the reference point. The extracted data was further partitioned into training and test sets based on the timestamp, with admissions prior to the year 2179 used for training and those afterward for testing to avoid data leakage during evaluation.

The data preprocessing resulted in a train set of 1,417,738 rows used for training and an independent test set of 100.000 rows. Each dataset contains 3 Columns representing the patient ID, patient admission ID, and Patient's race, as patient indicators. The dataset also contains 200 columns indicating the 100 lab values used and the time stamps, and other extra 200 columns for the follow-up values and timestamps. If the values are missing, those values where indicated using a Not a Number (NaN) value.

\subsection{Foundation Lab-MAE Architecture}

We base our foundation laboratory imputation model (Lab-MAE) on a Masked Autoencoder architecture, inspired by the Remasker framework \cite{du2023remasker}. This architecture utilizes a Transformer backbone to capture complex correlations between lab values over time, providing robust imputation of missing data in medical records. The model is composed of an encoder-decoder structure that is trained in a self-supervised manner by masking portions of the input and reconstructing the masked values.

The model uses learned positional encodings to represent the unique lab IDs, and timestamps ensuring that each lab value and timestamp is always passed to the model in the same positional slot in the input sequence. This approach allows the model to consistently interpret each lab test and time, regardless of missing values or the presence of other tests. Specifically, the lab values are placed in predefined positions, and the timestamps corresponding to those lab tests are placed in the following position. This design enables the model to capture temporal relationships between the lab values and their corresponding times. 

In this sense, let \( x \in \mathbb{R}^{L \times d} \) represent the input sequence, where \( L \) is the sequence length (including both lab values and timestamps) and \( d \) is the embedding dimension of each token. The learned positional encodings \( P \in \mathbb{R}^{L \times d} \) are added to the input sequence as follows:

\begin{equation} 
z_0 = x + P
\end{equation}

where \( z_0 \) is the input to the encoder. Positional encodings align lab values with corresponding timestamps, capturing temporal relationships.

Then, the encoder consists of multiple layers of Transformer blocks, where each block includes multi-head self-attention and feed-forward layers. The self-attention mechanism is defined as in the original "attention is all you need" paper \cite{vaswani2017attention}:

\begin{equation} 
\text{Attention}(Q, K, V) = \text{softmax}\left(\frac{QK^T}{\sqrt{d_k}}\right) V
\end{equation}

where \(Q\), \(K\), and \(V\) represent the query, key, and value matrices, respectively, and \(d_k\) is the dimensionality of the key vectors. To address the specific challenge of missing values in our dataset, we introduce a \textbf{missing value attention mask}, which prevents missing values from influencing the attention computation. Let \( M \in \{0, 1\}^{L \times L} \) represent the attention mask, where \(M_{ij} = 0\) indicates a missing value, and the corresponding attention score is masked out:

\begin{equation} 
    \tilde{A}_{ij} = 
    \begin{cases} 
    A_{ij}, & \text{if } M_{ij} = 1 \\
    -\infty, & \text{if } M_{ij} = 0 
    \end{cases}
\end{equation}

The decoder reconstructs the masked values by aggregating learnable masked tokens in the masked positions from the latent representation generated by the encoder. To ensure that missing values do not bias the model's predictions, we introduce modifications to the loss function. Specifically, the reconstruction loss is only calculated for observed values (current and masked values), while missing values are ignored. Given the predicted values \( \hat{x} \) and true values \( x \), the loss function \( L \) is defined as:

\begin{equation}
\label{eq:loss}
    L = \frac{1}{\sum_i (1 - m_i)} \sum_{i=1}^{L} (1 - m_i) \cdot (\hat{x}_i - x_i)^2
\end{equation}

where \( m_i \) is the missingness indicator for each value. This ensures that the model focuses on reconstructing only the available data, and prevents the overfitting to missing values.

\begin{figure*}[ht!]
    \centering
    \includegraphics[width=0.75\linewidth]{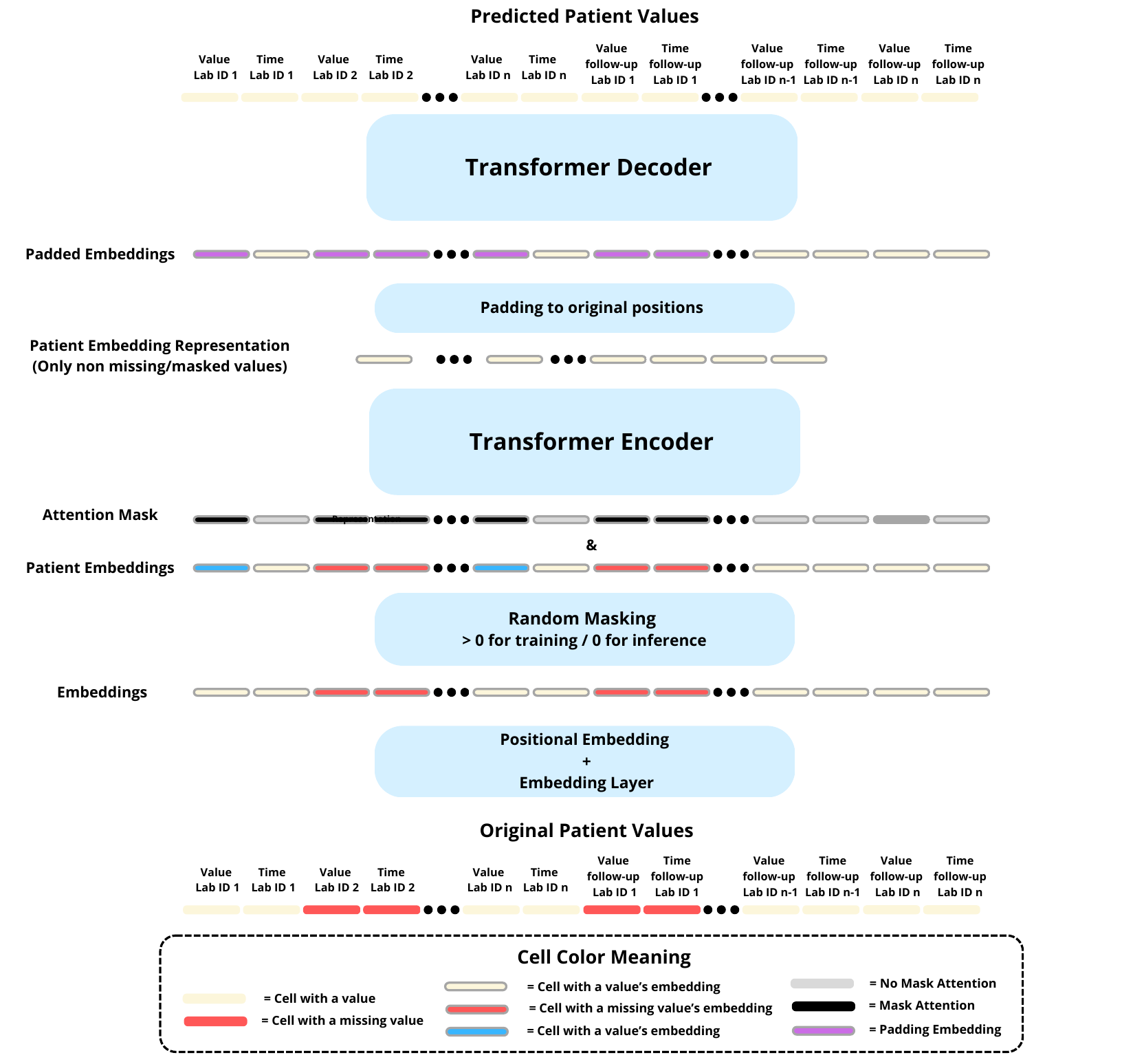}
    \caption{Foundation Lab-MAE training structure.}
    \label{fig:stream}
\end{figure*}

\subsection{Lab-MAE Training}

The Lab-MAE model was trained using a self-supervised learning approach with a focus on imputing missing values in clinical lab data. The training process involved several stages, including data preprocessing, model setup, optimization, and evaluation, designed to maximize the model's ability to predict missing lab values from the MIMIC-IV dataset \cite{johnson2020mimic, johnson2023mimic}.

Before training, the dataset underwent to a preprocessing to handle missing values and remove irrelevant or redundant data points. Rows with fewer than 17 non-missing lab and time values were excluded from the training set. The remaining data was normalized using a feature-wise min-max scaling technique to standardize the lab values across patients, enhancing the stability and convergence of the training process.

The Lab-MAE model was configured with the following hyperparameters: a hidden embedding dimension of 64, 8 layers for both the encoder and decoder, and 8 attention heads per layer. The model was trained with a mask ratio of 0.25, meaning that 25\% of the input values were randomly masked during training to simulate missing data. A batch size of 256 was used to ensure efficient training, and the training was carried out for a maximum of 500 epochs. A checkpoint mechanism was implemented to save model weights at regular intervals, allowing the model to be resumed from any epoch if necessary.

The optimization of the model was handled using the AdamW optimizer, which is well-suited for training Transformer-based architectures due to its effective weight decay mechanism. The learning rate was initially set using a base learning rate scaling rule, proportional to the batch size. During training, a cosine annealing schedule was employed to adjust the learning rate dynamically. Specifically, the learning rate was warmed up linearly over the first 20 epochs, after which it followed a half-cycle cosine decay pattern, gradually reducing towards a specified minimum learning rate. 




The model's training objective was to minimize the Mean Squared Error (MSE) between the predicted and actual lab values, calculated only over the non-masked (observed) entries as shown in the equation \ref{eq:loss}.

The model was evaluated constantly on a validation set extracted from the training set. The validation was performed at regular intervals, with predictions assessed on standard additional metrics such as Root Mean Squared Error (RMSE), R-squared (R²), Mean Absolute Error (MAE), and Wasserstein Distance (WD). The validation results were logged after every 30 epochs to monitor the model's performance. Additionally, checkpointing enabled the continuous saving of the model's state, allowing for the resumption or rollback of training to optimize performance based on the evaluation metrics.

This structured training approach enabled the model to learn robust representations of the temporal and contextual relationships between lab values, ultimately enhancing its ability to accurately impute missing values in the dataset and learn representations of the lab values.

\subsection{Lab-MAE Imputation Evaluation}

\subsubsection{Baseline Models}

To benchmark the performance of our proposed Lab-MAE model for lab value imputation, we implemented a set of baseline models using XGBoost, softimpute, GAIN, EM, and MICE. For XGBoost implementation, a total of 100 separate XGBoost models, were trained, one for each lab value. For softimpute, GAIN, EM, and MICE, the models were implemented using hyperimpute \cite{hyperimpute}.

The training of these XGBoost models involved a hyperparameter optimization process using GridSearchCV to identify the best-performing configuration for each lab-specific model. The hyperparameter search was conducted using the following grid: Learning rate: [0.01, 0.1, 0.2], Max depth: [3, 4, 5], Number of estimators: [50, 100, 200].

GridSearchCV was applied with three-fold cross-validation to evaluate different parameter combinations, aiming to minimize the MSE of the lab value of interest. After identifying the optimal hyperparameters, each model was retrained using the entire training dataset to ensure the best possible predictive performance.

\subsubsection{Lab-MAE Imputation Evaluation Setup}

Both the Lab-MAE and baseline models (XGBoost, softimpute, GAIN, EM, and MICE) were evaluated using a cohort of 100,000 data points extracted from our independent test set. To ensure a fair comparison, the same data points were used across both models during evaluation. We computed three primary metrics for each lab value: RMSE, R\(^2\), and the WD. The R\(^2\) metric was used as the main reference due to its ability to assess the correlation strength and its capacity to avoid overfitting to mean or most common values. The WD was used to assess how well the model captures the overall distribution of lab values, including extreme values, unlike RMSE and R\(^2\), which emphasize overall accuracy and variance explained. WD complements RMSE and R\(^2\) by highlighting the quality of predictions in capturing not only central tendencies but also the full range of values, particularly the extremes. More information about the metrics is avaiable in appendix \ref{apd:metrics}.

\subsubsection{Test Set Evaluation Process}

The evaluation was conducted lab-by-lab to ensure comprehensive and robust performance assessment. For each lab value, we simulated missing data by masking its existing values in the test dataset, effectively challenging the models to predict these masked values using the remaining available context, including other lab values and the associated timestamps.

During inference with the Lab-MAE model, we ensured that no gradients were calculated, and the random masking ratio was set to zero, focusing purely on the prediction task.

Similarly, the baseline methods were trained in a lab-specific manner, where each lab's missing values were predicted using each imputation method.

\subsection{Fairness and Bias Analysis}

The fairness and bias analysis in our study aimed to evaluate how well the Lab-MAE model performs across different demographic groups, particularly focusing on racial differences, and to assess the impact of follow-up data as a potential shortcut feature in the imputation process.

\subsubsection{Lab-MAE Model Fairness Across Race Groups}

To ensure that the Lab-MAE model's predictions are equitable across different racial groups, we conducted an analysis of its performance for five race groups: White, Black, Hispanic, Asian, and Others. We compared the performance of the Lab-MAE model with the best-performed baseline model (XGBoost) for each race group using the same metrics.

For each race, the imputation model's performance was calculated by comparing the predicted values to the actual lab values, using the following approach:

\begin{itemize}
    \item Filter the test dataset to include only the records for the race group being evaluated.
    \item Mask the lab values to simulate missing data and use the Lab-MAE model to predict these values.
    \item Calculate the metrics WD, RMSE, and R\(^2\) for each lab value to quantify the model's prediction performance for the respective race group.
\end{itemize}

The results were consolidated into a single dataframe to allow a comparative analysis of the model's fairness across racial categories.

\subsubsection{Carbon Footprint Measurement}

We assessed the carbon footprint of the Lab-MAE and the best-performed baseline (XGBoost), using the CodeCarbon library \cite{benoit_courty_2024_14518377}. Emissions were estimated during the inference process for batch sizes of 1, 32, and 64 across multiple geographic locations, including Colombia, USA, France, Uganda, Philippines, and Australia. These locations were chosen to represent diverse geographic profiles in Asia, Africa, North America, South America, Europe, and Australia.

The models were evaluated by performing inference on subsets on the test dataset comparing the performance of the Lab-MAE model vs the set of XGBoost models. For each location, emissions were measured using simulated conditions based on the country’s emission factor, reflecting the environmental impact of running machine learning models in different parts of the world.

More details about the methodology used to calculate carbon emissions can be found in Appendix \ref{apd:carbon}. 

\section{Results}
\subsection{Lab-MAE imputation performance}

In this section, we present the results of our comparison between the Lab-MAE and the baseline models (XGBoost, softimpute, GAIN, EM, and MICE) for the task of data imputation. We evaluated using three metrics: WD, RMSE, and R\(^2\). These metrics were calculated on the entire set of 100 lab values. A further detailed comparison with XGBoost on the top 20 most frequently occurring lab values  was performed to provide more insights about model performance.

\subsubsection{Overall Analysis of All Lab Values}

To mitigate biases arising from differences in the scale of lab measurements and the inherent imbalance in lab value frequencies, we aggregated the evaluation metrics across 100 lab tests by counting the number of tests in which each baseline method outperforms Lab-MAE. In this analysis, for RMSE and WD, lower values indicate superior performance, whereas for R\(^2\) higher values are preferable. This aggregation strategy provides a more balanced assessment of performance across heterogeneous lab tests.

\begin{table}[ht]
    \centering
    \caption{Comparison of Lab-MAE with baseline imputation methods across evaluation metrics. The figures represent the number of lab tests (out of 100) where the baseline method outperforms Lab-MAE. For RMSE and WD, lower values indicate better performance, whereas for R\(^2\), higher values are preferred.}
    \scriptsize
    \begin{tabular}{lccc}
        \toprule
        \textbf{Model vs Ours} & \textbf{RMSE} & \textbf{R\(^2\)} & \textbf{Wasserstein Distance} \\
        \midrule
        XGBoost   & 11\%  & 11\%  & 21\% \\
        softimpute & 14\%  & 0\%   & 6\%  \\
        GAIN      & 3\%   & 5\%   & 11\% \\
        EM        & 4\%   & 12\%  & 12\% \\
        MICE      & 11\%  & 18\%  & 16\% \\
        \bottomrule
    \end{tabular}
    \label{tab:metric-comparison}
\end{table}

Table \ref{tab:metric-comparison} summarizes the comparison between Lab-MAE and the baseline imputation methods—XGBoost, softimpute, GAIN, EM, and MICE. For example, when compared to XGBoost, Lab-MAE achieves lower RMSE and higher R\(^2\) in 89 out of 100 lab tests, and a lower WD in 79 tests. In contrast, the other baselines outperform Lab-MAE in fewer tests: softimpute surpasses Lab-MAE in 14 tests for RMSE, 6 for WD, and in none for R\(^2\); GAIN outperforms in 3 (RMSE), 11 (WD), and 5 (R\(^2\)) tests; EM does so in 4 (RMSE), 12 (WD), and 12 (R\(^2\)) tests; and MICE in 11 (RMSE), 16 (WD), and 18 (R\(^2\)) tests.

\subsubsection{Focused Analysis on the Top 20 Lab Values}

To provide a clearer understanding of the models' performance on the most relevant lab values, we analyzed the top 20 most frequently occurring lab values. This focused analysis highlights the differences in lab value imputaion between Lab-MAE and the best-performed model-XGBoost.

\begin{table*}[ht]
    \centering
    \scriptsize
    \caption{Comparison of RMSE, WD, and R\(^2\) between XGBoost and Lab-MAE models for top 20 most popular lab values.}
    \scriptsize
    \resizebox{\textwidth}{!}{
    \begin{tabular}{lcccccc}
        \toprule
        Lab ID & RMSE (XGBoost) & RMSE (Lab-MAE) & WD (XGBoost) & WD (Lab-WD) & R\(^2\) (XGBoost) & R\(^2\) (Lab-MAE) \\
        \midrule
        Creatinine & 0.261 & \textbf{0.233} & 0.039 & \textbf{0.034} & 0.929 & \textbf{0.943} \\
        Hematocrit & 0.579 & \textbf{0.564} & 0.046 & \textbf{0.040} & 0.989 & 0.989 \\
        Potassium & 0.327 & \textbf{0.270} & 0.127 & \textbf{0.095} & 0.494 & \textbf{0.654} \\
        Sodium & 1.035 & \textbf{0.883} & 0.309 & \textbf{0.220} & 0.921 & \textbf{0.943} \\
        Urea Nitrogen & 4.910 & \textbf{4.531} & 0.750 & \textbf{0.540} & 0.913 & \textbf{0.925} \\
        Chloride & 1.049 & \textbf{0.876} & 0.310 & \textbf{0.229} & 0.952 & \textbf{0.967} \\
        Bicarbonate & 0.959 & \textbf{0.769} & 0.305 & \textbf{0.199} & 0.927 & \textbf{0.953} \\
        Anion Gap & 1.008 & \textbf{0.808} & 0.356 & \textbf{0.208} & 0.872 & \textbf{0.918} \\
        Platelet Count & 41.464 & \textbf{39.861} & 9.349 & \textbf{7.568} & 0.855 & \textbf{0.866} \\
        Hemoglobin & 0.093 & \textbf{0.080} & 0.025 & \textbf{0.026} & 0.998 & 0.998 \\
        White Blood Cells & 2.034 & \textbf{1.935} & 0.551 & \textbf{0.461} & 0.757 & \textbf{0.780} \\
        MCHC & 0.194 & \textbf{0.114} & 0.035 & \textbf{0.027} & 0.982 & \textbf{0.994} \\
        Red Blood Cells & 0.044 & \textbf{0.035} & 0.005 & \textbf{0.004} & 0.996 & \textbf{0.997} \\
        MCV & 0.518 & \textbf{0.401} & 0.243 & \textbf{0.238} & 0.993 & \textbf{0.996} \\
        MCH & 0.159 & \textbf{0.109} & 0.030 & \textbf{0.029} & 0.995 & \textbf{0.998} \\
        RDW & 0.579 & \textbf{0.549} & 0.084 & \textbf{0.073} & 0.931 & \textbf{0.938} \\
        Glucose & 30.392 & \textbf{29.277} & 11.667 & \textbf{10.095} & 0.426 & \textbf{0.468} \\
        Magnesium & 0.184 & \textbf{0.178} & 0.079 & \textbf{0.071} & 0.457 & \textbf{0.491} \\
        Calcium, Total & 0.342 & \textbf{0.324} & 0.097 & \textbf{0.082} & 0.691 & \textbf{0.723} \\
        Phosphate & 0.536 & \textbf{0.511} & 0.187 & \textbf{0.155} & 0.615 & \textbf{0.650} \\
        \bottomrule
    \end{tabular}
    }
    \label{tab:top20-comparison}
\end{table*}

Table \ref{tab:top20-comparison} illustrates the detailed performance metrics for both models on the top 20 most popular lab values. The Lab-MAE model consistently demonstrates superior performance, achieving lower RMSE and WD values, as well as higher R\(^2\) scores for most lab values.

For example, for Creatinine, the Lab-MAE model achieved an RMSE of 0.233 compared to 0.261 for XGBoost, indicating a significant reduction in the average error. Furthermore, the WD for Lab-MAE was 0.034, which is lower than the 0.039 obtained by XGBoost. The R\(^2\) score for Lab-MAE was 0.943, outperforming XGBoost's 0.929, demonstrating a stronger predictive capability.

This trend is not isolated to Creatinine; similar patterns are observed across other lab values. For Sodium, Lab-MAE showed a reduction in RMSE from 1.035 to 0.883 and in WD from 0.309 to 0.220, with an improvement in R\(^2\) from 0.921 to 0.943, highlighting the model's robustness in reducing both large and small prediction errors.

In some cases, such as lab Platelet Count, both models presented challenges due to the complexity of the predictions. Nevertheless, Lab-MAE still showed slight improvements, with an RMSE of 39.861 versus 41.464 for XGBoost and an WD of 7.568 compared to 9.349. The R\(^2\) values for this lab remained relatively high for both models, with Lab-MAE marginally outperforming XGBoost (0.866 versus 0.855).

The detailed comparison provided in Table \ref{tab:top20-comparison} underscores the strengths of the Lab-MAE model, suggesting that its ability to leverage the temporal and contextual information encoded in the dataset is a significant advantage over the XGBoost approach. These findings validate the hypothesis that a Transformer-based architecture, when properly trained and fine-tuned, can substantially improve data imputation tasks over classical machine learning methods.

The analysis confirms that the Lab-MAE model not only performs better overall but also excels particularly in the top 20 most frequently ordered lab values, making it a superior choice for data imputation tasks in clinical settings.

\subsection{Fairness Analysis Across Race Groups}

To gain deeper insights into the bias and fairness of the Lab-MAE and XGBoost models, we conducted a focused analysis on the top 20 most frequently occurring lab values, evaluating their performance across different racial groups. This detailed analysis reveals notable patterns in how each model performs for specific lab values, highlighting both strengths and potential biases.

Table \ref{tab:top20_race_comparison}, \ref{tab:top20_race_comparison_wd}, and \ref{tab:top20_race_comparison_rmse} available in Appendix \ref{apd:metrics-race}, illustrate the \(R^2\), WD, and RMSE values respectively of the Lab-MAE model across different racial groups for the top 20 lab values. The analysis reveals that the model's performance varies significantly depending on both the lab value and the racial group, indicating potential biases that warrant further attention.

One notable pattern is the consistent higher performance of the Lab-MAE model for Asian and White groups in certain lab values, which indicates a disparity in the model performance. This disparity in the performance is dependent on the lab value and metric used, highlighting the need of multiple metrics and domain specific metrics.

\subsection{Carbon Footprint Results}
After running experiments on inference of the Lab-MAE model versus the XGBoost models in batches of 1, 32 and 64 data points, and collecting the results, we averaged across six geographic locations spanning Asia, Africa, Australia, Europe, North America, and South America. Table~\ref{tab:carbon_footprint} shows the mean values of duration, emissions, emissions rate, CPU power, GPU power, and RAM power for each model and batch size. Overall, Lab-MAE shows lower or comparable carbon footprints, particularly at lower batch sizes, whereas XGBoost sometimes demands a higher CPU power usage given the need of requiring a model per feature. These results provide insight into the energy efficiency of each model under different workload configurations.

\begin{table*}[ht!]
\scriptsize
\centering
\caption{Carbon footprint of Lab-MAE and XGBoost under different batch sizes. Values represent the \emph{average calculation using the mean} of the six geographic locations in South America, North America, Europe, Asia, Africa, and Australia.}
\label{tab:carbon_footprint}
\begin{tabular}{cccccccc}
\toprule
Batch & Model & Duration & Emissions & Emissions & CPU & GPU & RAM \\
Size &  & (s) & kg CO\(_2\) & Rate & Power (W) & Power (W) & Power (W) \\
\midrule
\multirow{2}{*}{1} & Lab-MAE     & \textbf{6.415688}  & \textbf{4.251868e-08} & \textbf{6.657141e-09} & 0.50625   & 0.028650  & 3.0 \\
                   & XGBoost & 10.866340 & 1.363608e-06 & 1.242559e-07 & \textbf{0.21970}   & \textbf{0.033817}  & 3.0 \\
\midrule
\multirow{2}{*}{32} & Lab-MAE     & \textbf{6.993714}  & \textbf{2.271253e-07} & \textbf{3.237850e-08} & 0.257067  & 0.021300  & 3.0 \\
                    & XGBoost & 11.703058 & 2.935091e-06 & 2.389255e-07 & \textbf{1.908717}  & \textbf{0.046033}  & 3.0 \\
\midrule
\multirow{2}{*}{64} & Lab-MAE     & \textbf{7.826687}  & \textbf{5.551631e-07} & \textbf{7.066615e-08} & 0.722167  & 0.041750  & 3.0 \\
                    & XGBoost & 10.946252 & 1.403389e-06 & 1.267618e-07 & \textbf{0.278117}  & \textbf{0.035150}  & 3.0 \\
\bottomrule
\end{tabular}
\end{table*}

As shown, Lab-MAE tends to have a shorter average duration (especially with smaller batch sizes), resulting in generally lower carbon emissions. XGBoost, although sometimes requiring lower CPU power, demonstrates higher total duration in the same scenarios, which contributes to a higher overall emissions rate. The duration is mainly because Lab-MAE is a single model, while XGBoost requires a model per lab value. These observations suggest that Lab-MAE may be more energy-efficient when deployed at scale since is a single foundation model.

\section{Discussion}

Our study introduces Lab-MAE, a novel transformer-based architecture that fundamentally advances the field of clinical data imputation while maintaining algorithmic fairness. Through an extensive empirical evaluation of the MIMIC-IV dataset \cite{johnson2020mimic, johnson2023mimic}, we demonstrate that Lab-MAE achieves superior performance compared to traditional approaches and exhibits remarkable consistency between demographic groups - a critical consideration for healthcare applications.

\subsection{Clinical and Technical Implications}

Our LAB-MAE demonstrates a greater ability to learn real-world distributions, which may be characterized by extreme values. This is highlighted by consistent R2 above baselines but reduced EMD/Wasserstein values overall and across subgroups. 

The model's performance stability across demographic groups challenges the expected trade-off between accuracy and fairness in machine learning systems. This suggests that architectural innovations focused on capturing temporal and contextual relationships can simultaneously advance both objectives.


\subsection{Fairness and Equity Considerations}

Previous approaches to laboratory value imputation have largely relied on traditional machine learning methods or simplified time-series models. Our model builds upon recent work by \cite{bellamy2023labrador} for modeling laboratory data, and \cite{du2023remasker} with the ReMasker framework for tabular data imputation, while addressing their limitations, particularly in handling temporal dependencies and maintaining performance across diverse patient populations. The significant improvement over the baseline models, especially in complex laboratory parameters, aligns with emerging evidence that properly architected deep learning models can overcome the traditional advantages of tree-based methods in tabular data \cite{van2024tabular}.  However, the observed disparities in performance across racial groups reflect the social patterning of data generation \cite{teotia2024variation}, where factors such as systemic inequalities in healthcare access contribute to missingness patterns. Addressing this requires not only technical innovations in imputation models but also systemic efforts to improve data equity

\subsection{Connection to Foundation Models}
The success of Lab-MAE reflects a broader paradigm shift in medical AI, where foundation model architectures are being successfully adapted to specialized clinical tasks. Similar to how large language models have revolutionized natural language processing, our results suggest that transformer-based architectures can effectively capture clinical data's complex temporal and interdependent nature. The robust performance of Lab-MAE across diverse patient populations suggests that foundation model architectures when properly adapted to clinical domains, can help bridge the gap between general and specialized medical AI applications.

By consolidating multiple tasks into a single, cohesive model, Lab-MAE reduces the redundancy and overhead associated with training and deploying separate models for each feature, as required by XGBoost or MICS. This capability is particularly advantageous in healthcare settings like hospitals, where complex data environments demand robust yet streamlined solutions. Lab-MAE’s foundation model design highlights its potential to adapt effectively to diverse clinical scenarios, offering a powerful combination of scalability and environmental responsibility.

\subsection{Limitations and Future Directions}
Despite Lab-MAE's promising results, several limitations merit attention. First, our evaluation was conducted on a single, academic medical center dataset, potentially limiting generalizability. Second, while we demonstrated fairness across major demographic groups, future work should investigate intersectional fairness and rare subpopulations. Key directions for future research include: (1) extending Lab-MAE to incorporate structured medical knowledge, (2) investigating transfer learning capabilities across different healthcare settings, and (3) developing interpretability methods specifically designed for temporal clinical predictions.

\section{Conclusion}
Lab-MAE represents a significant advance in clinical data imputation, demonstrating that foundation model architectures can be effectively adapted for specialized healthcare tasks while maintaining fairness across demographic groups. Our results suggest a promising path forward for developing robust, equitable healthcare AI systems that can handle the complexity of real-world clinical data. As healthcare continues to digitize and generate increasingly complex datasets, approaches like Lab-MAE will be crucial for ensuring both high performance and algorithmic fairness in clinical decision support systems.

\newpage
\appendix

\appendix

\clearpage
\section{Performance metrics}\label{apd:metrics}
\textbf{Evaluation Metrics:} The following metrics were used to measure the accuracy of the predictions for each lab value:

\begin{itemize}

    \item \textbf{RMSE}: Represents the square root of the average squared differences between the predicted and actual values, emphasizing larger errors.
    \begin{equation}
    \text{RMSE} = \sqrt{\frac{1}{n} \sum_{i=1}^{n} (y_i - \hat{y}_i)^2}
    \end{equation}
    where \( n \) is the number of data points, \( y_i \) represents the true values, and \( \hat{y}_i \) represents the predicted values.

    \item \textbf{R\(^2\)}: Evaluates the proportion of variance in the dependent variable that can be explained by the model, with values closer to 1 indicating a stronger predictive capability.
    \begin{equation}
    R^2 = 1 - \frac{\sum_{i=1}^{n} (y_i - \hat{y}_i)^2}{\sum_{i=1}^{n} (y_i - \bar{y})^2}
    \end{equation}
    where \( \bar{y} \) is the mean of the observed data, \( y_i \) represents the true values, and \( \hat{y}_i \) represents the predicted values.

    \item \textbf{Wasserstein Distance}: WD measures the distance between two probability distributions \cite{villani2009wasserstein}. It is particularly sensitive to differences in the tails of distributions, making it a suitable for capturing the ability of the model to predict abnormal values \cite{panaretos2019statistical}. Given two distributions \( P \) (actual values) and \( Q \) (predicted values), the 1st Wasserstein distance is defined as:
    \begin{equation}
    W(P, Q) = \inf_{\gamma \in \Gamma(P, Q)} \int_{\mathbb{R}^2} \lVert x - y \rVert \, d\gamma(x, y)
    \end{equation}
    where \( \Gamma(P, Q) \) represents the set of all joint distributions with marginals \( P \) and \( Q \). In our implementation, WD is approximated using the cumulative distribution functions (CDFs) of the predicted and actual values:
    \begin{equation}
    W(P, Q) \approx \int_{-\infty}^{\infty} \lvert F_P(x) - F_Q(x) \rvert dx
    \end{equation}
    where \( F_P(x) \) and \( F_Q(x) \) are the CDFs of the actual and predicted distributions, respectively.
\end{itemize}

This systematic evaluation allowed for a direct comparison of the Lab-MAE and XGBoost models, highlighting their relative strengths and weaknesses in accurately imputing missing lab values in the medical dataset. The consistent use of the same test data points for both models ensured a fair and unbiased evaluation of their performance in this critical imputation task.

\section{Carbon footprint measurement}\label{apd:carbon}

\subsection{Methods: Carbon footprint measurement}

The total emissions \( E \) were calculated using \cite{benoit_courty_2024_14518377} as:

\begin{equation}
E = \int_0^T P(t) \, dt
\end{equation}

where \( P(t) \) is the power consumption at time \( t \), and \( T \) is the total runtime. Power consumption was derived as:

\begin{equation}
P(t) = P_{\text{CPU}}(t) + P_{\text{GPU}}(t) + P_{\text{RAM}}(t)
\end{equation}

The energy consumption \( \text{E}_{\text{consumed}} \) is calculated as:

\begin{equation}
\text{E}_{\text{consumed}} = \sum_{i=1}^{n} \left( P_{\text{CPU}} \times t_i + P_{\text{GPU}} \times t_i + P_{\text{RAM}} \times t_i \right)
\end{equation}

Carbon emissions \( \text{Emissions} \) are then computed as:

\begin{equation}
\text{Emissions} = \text{E}_{\text{consumed}} \times \text{EmissionFactor}
\end{equation}

where \( \text{EmissionFactor} \) depends on the energy grid of the location, accounting for the carbon intensity of electricity. The information of the EmissionFactor per country is available in \cite{codecarbon_global_energy_mix}.

\subsection{Results: Carbon footprint measurement}

\begin{table*}[ht!]
    \centering
    \caption{Carbon Emissions and Power Usage for Batch Size 1}
    \small
    \resizebox{\textwidth}{!}{
    \begin{tabular}{lcccccc}
        \toprule
        Continent & Model & Duration (s) & Emissions (kg CO\(_2\)) & Emissions Rate & CPU Power (W) & GPU Power (W) \\
        \midrule
        Africa      & Lab-MAE & 6.35 & \(4.85 \times 10^{-9}\) & \(7.63 \times 10^{-10}\) & 0.22 & 0.00 \\
        Africa      & XGBoost     & 10.67 & \(1.82 \times 10^{-7}\) & \(1.71 \times 10^{-8}\) & 0.32 & 0.00 \\
        Asia        & Lab-MAE & 6.33 & \(4.92 \times 10^{-8}\) & \(7.77 \times 10^{-9}\) & 0.12 & 0.00 \\
        Asia        & XGBoost     & 10.82 & \(2.63 \times 10^{-6}\) & \(2.43 \times 10^{-7}\) & 0.29 & 0.05 \\
        Australia   & Lab-MAE & 6.41 & \(1.05 \times 10^{-7}\) & \(1.63 \times 10^{-8}\) & 0.63 & 0.10 \\
        Australia   & XGBoost     & 11.28 & \(2.54 \times 10^{-6}\) & \(2.25 \times 10^{-7}\) & 0.33 & 0.11 \\
        Europe      & Lab-MAE & 6.66 & \(8.27 \times 10^{-9}\) & \(1.24 \times 10^{-9}\) & 0.26 & 0.07 \\
        Europe      & XGBoost     & 10.74 & \(2.15 \times 10^{-7}\) & \(2.00 \times 10^{-8}\) & 0.06 & 0.00 \\
        North America & Lab-MAE & 6.41 & \(4.23 \times 10^{-8}\) & \(6.59 \times 10^{-9}\) & 0.50 & 0.00 \\
        North America & XGBoost     & 11.01 & \(1.64 \times 10^{-6}\) & \(1.48 \times 10^{-7}\) & 0.27 & 0.05 \\
        South America & Lab-MAE & 6.33 & \(4.59 \times 10^{-8}\) & \(7.26 \times 10^{-9}\) & 1.31 & 0.00 \\
        South America & XGBoost     & 10.68 & \(9.82 \times 10^{-7}\) & \(9.20 \times 10^{-8}\) & 0.05 & 0.00 \\
        \bottomrule
    \end{tabular}
    }
    \label{tab:batch1}
\end{table*}

\begin{table*}[ht!]
    \centering
    \caption{Carbon Emissions and Power Usage for Batch Size 32}
    \small
    \resizebox{\textwidth}{!}{
    \begin{tabular}{lcccccc}
        \toprule
        Continent & Model & Duration (s) & Emissions (kg CO\(_2\)) & Emissions Rate & CPU Power (W) & GPU Power (W) \\
        \midrule
        Africa      & Lab-MAE & 7.02 & \(2.96 \times 10^{-8}\) & \(4.22 \times 10^{-9}\) & 0.07 & 0.02 \\
        Africa      & XGBoost     & 10.95 & \(1.82 \times 10^{-7}\) & \(1.66 \times 10^{-8}\) & 0.10 & 0.02 \\
        Asia        & Lab-MAE & 6.94 & \(3.77 \times 10^{-7}\) & \(5.44 \times 10^{-8}\) & 0.06 & 0.00 \\
        Asia        & XGBoost     & 12.76 & \(1.10 \times 10^{-5}\) & \(8.62 \times 10^{-7}\) & 8.37 & 0.14 \\
        Australia   & Lab-MAE & 7.09 & \(4.35 \times 10^{-7}\) & \(6.14 \times 10^{-8}\) & 0.15 & 0.10 \\
        Australia   & XGBoost     & 11.95 & \(3.17 \times 10^{-6}\) & \(2.66 \times 10^{-7}\) & 0.60 & 0.10 \\
        Europe      & Lab-MAE & 6.93 & \(3.55 \times 10^{-8}\) & \(5.13 \times 10^{-9}\) & 0.17 & 0.00 \\
        Europe      & XGBoost     & 12.42 & \(4.30 \times 10^{-7}\) & \(3.46 \times 10^{-8}\) & 1.67 & 0.01 \\
        North America & Lab-MAE & 7.06 & \(3.24 \times 10^{-7}\) & \(4.59 \times 10^{-8}\) & 1.04 & 0.01 \\
        North America & XGBoost     & 11.22 & \(1.64 \times 10^{-6}\) & \(1.46 \times 10^{-7}\) & 0.24 & 0.00 \\
        South America & Lab-MAE & 6.94 & \(1.61 \times 10^{-7}\) & \(2.32 \times 10^{-8}\) & 0.06 & 0.00 \\
        South America & XGBoost     & 10.92 & \(1.19 \times 10^{-6}\) & \(1.09 \times 10^{-7}\) & 0.47 & 0.00 \\
        \bottomrule
    \end{tabular}
    }
    \label{tab:batch32}
\end{table*}

\begin{table}[ht!]
    \centering
    \caption{Carbon Emissions and Power Usage for Batch Size 64}
    \small
    \resizebox{\textwidth}{!}{
    \begin{tabular}{lcccccc}
        \toprule
        Continent & Model & Duration (s) & Emissions (kg CO\(_2\)) & Emissions Rate & CPU Power (W) & GPU Power (W) \\
        \midrule
        Africa      & Lab-MAE & 7.85 & \(6.33 \times 10^{-8}\) & \(8.07 \times 10^{-9}\) & 0.13 & 0.00 \\
        Africa      & XGBoost     & 10.70 & \(1.92 \times 10^{-7}\) & \(1.79 \times 10^{-8}\) & 0.45 & 0.00 \\
        Asia        & Lab-MAE & 7.87 & \(1.23 \times 10^{-6}\) & \(1.56 \times 10^{-7}\) & 1.07 & 0.10 \\
        Asia        & XGBoost     & 11.20 & \(2.94 \times 10^{-6}\) & \(2.63 \times 10^{-7}\) & 0.36 & 0.10 \\
        Australia   & Lab-MAE & 7.88 & \(8.91 \times 10^{-7}\) & \(1.13 \times 10^{-7}\) & 0.35 & 0.10 \\
        Australia   & XGBoost     & 11.07 & \(2.45 \times 10^{-6}\) & \(2.21 \times 10^{-7}\) & 0.20 & 0.10 \\
        Europe      & Lab-MAE & 7.70 & \(7.82 \times 10^{-8}\) & \(1.02 \times 10^{-8}\) & 0.24 & 0.04 \\
        Europe      & XGBoost     & 10.76 & \(2.34 \times 10^{-7}\) & \(2.18 \times 10^{-8}\) & 0.31 & 0.00 \\
        North America & Lab-MAE & 7.83 & \(4.99 \times 10^{-7}\) & \(6.38 \times 10^{-8}\) & 0.10 & 0.01 \\
        North America & XGBoost     & 11.03 & \(1.52 \times 10^{-6}\) & \(1.38 \times 10^{-7}\) & 0.13 & 0.00 \\
        South America & Lab-MAE & 7.83 & \(5.73 \times 10^{-7}\) & \(7.32 \times 10^{-8}\) & 2.44 & 0.00 \\
        South America & XGBoost     & 10.93 & \(1.08 \times 10^{-6}\) & \(9.85 \times 10^{-8}\) & 0.21 & 0.00 \\
        \bottomrule
    \end{tabular}
    }
    \label{tab:batch64}
\end{table}

\clearpage
\section{Impact of shortcut features}\label{apd:first}

\subsection{Methods: Impact of follow-up data as a shortcut feature}

In addition to analyzing fairness across races, we also examined the Lab-MAE model's performance in scenarios where follow-up data was present versus when it was absent. This analysis aimed to understand if the availability of follow-up data serves as a shortcut feature, potentially inflating the model's performance by providing additional context.

For this analysis, the following procedure was adopted:
\begin{itemize}
    \item For each lab value, we identified the corresponding follow-up values (denoted as \texttt{npval\_last}) and separated the test samples into two groups: those with follow-up values and those without.
    \item The imputation model's performance was then evaluated separately for each group to determine how the presence or absence of follow-up data affected its predictive performance of the model on data imputation.
    \item Metrics such as WD, and R\(^2\) were calculated for each group to quantify the performance variations between the two scenarios.
\end{itemize}

This experiment allowed us to observe whether the model's predictive performance disproportionately relies on the availability of follow-up data, which could lead to biased predictions when such data is not present.

The findings from both the fairness analysis across racial groups and the follow-up data evaluation are crucial for understanding the Lab-MAE model's robustness and its potential biases. This analysis also aids in identifying areas where the model could be further improved to ensure more equitable performance across diverse patient populations.

\subsection{Results: Impact of follow-up data as a shortcut feature}

In this section, we analyze the impact of the Lab-MAE and XGBoost models to shortcuts by comparing their imputation performance with and without the presence of follow-up values. Follow-up values represent additional lab measurements taken after the initial test, providing a temporal context that could serve as a shortcut for predicting the target lab values.

To do so, we evaluated the performance of both models using the R\(^2\) metrics for the lab values with and without follow-up data. The comparison of the distributions for each lab value and the WD values is available in the supplementary materials.

\begin{table}[ht]
    \centering
    \caption{Comparison of Lab-MAE and XGBoost models with and without follow-up values for imputation.}
    \begin{tabular}{lcccc}
        \toprule
        Model & Follow-Up & R² \\
        \midrule
        XGBoost & \checkmark & 0.7391 \\
                 &  & 0.6089 \\
        Lab-MAE & \checkmark & 0.7661 \\
                 &  & 0.6544 \\
        \bottomrule
    \end{tabular}
    \label{tab:follow-up-comparison}
\end{table}

As illustrated in Table \ref{tab:follow-up-comparison}, both models experience a noticeable decline in performance when follow-up values are absent. The \(R^2\) values reflect a decrease in model performance without follow-up values. For XGBoost, \(R^2\) drops from 0.7391 with follow-up to 0.6089 without follow-up. Lab-MAE shows a similar pattern, with \(R^2\) declining from 0.7661 with follow-up to 0.6544 without. These results suggest that both models benefit significantly from the additional temporal information provided by follow-up values.

The Lab-MAE model outperforms XGBoost across both scenarios, demonstrating a smaller reduction in accuracy when follow-up values are excluded. This indicates that the Lab-MAE model is more robust in handling cases where only the initial lab value is available, further highlighting the effectiveness of its Transformer-based architecture in capturing complex patterns even with limited data.

\clearpage
\section{R2 metrics comparison}\label{apd:second}

\subsection{R2 metrics comparison of Lab-MAE vs XGBoost for the top 20 lab values}
\begin{figure}[ht!]
    \centering
    \resizebox{\textwidth}{!}{
    \includegraphics[width=2\linewidth]{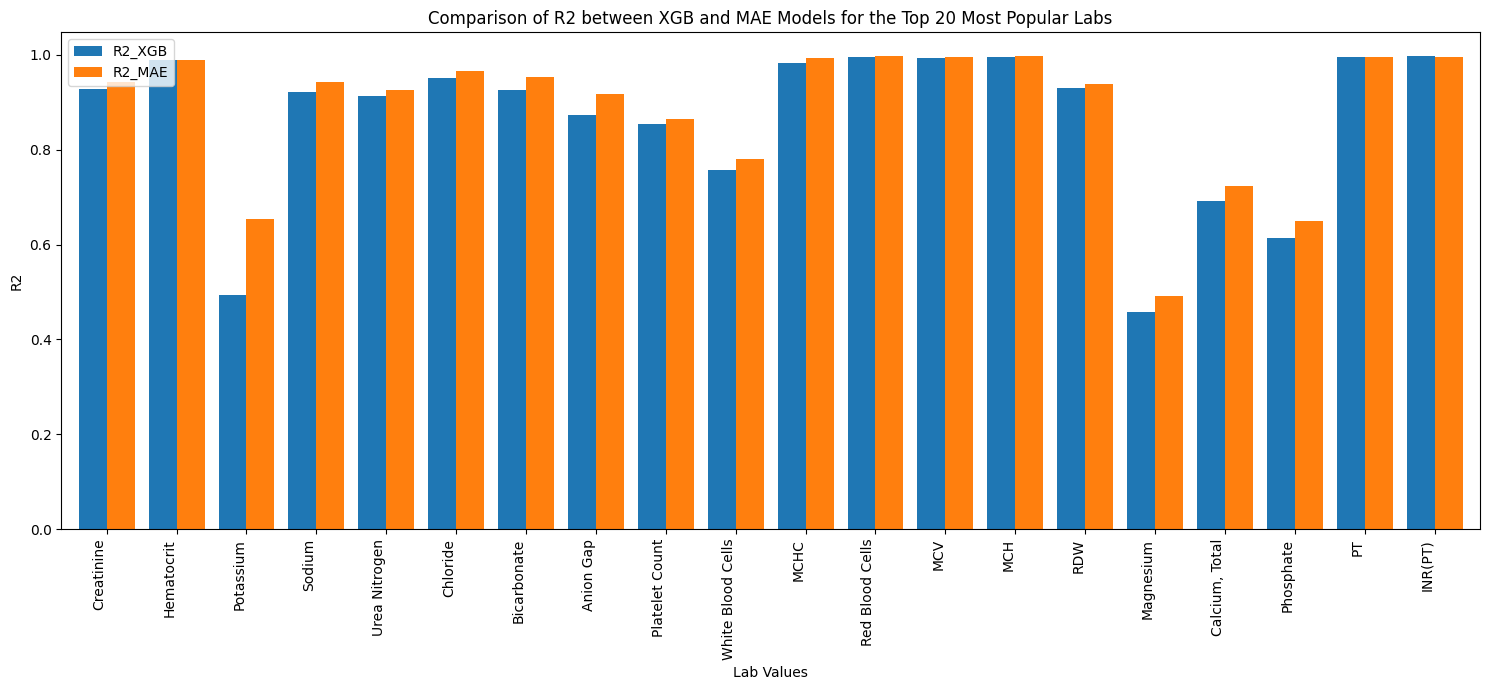}
    }
    \caption{Comparison of R\(^2\) between Lab-MAE and XGBoost models for the top 20 lab values. Orange is our Lab-MAE model and blue is XGBoost}
    \label{fig:top20-comparison}
\end{figure}

Figure \ref{fig:top20-comparison} provides a visual representation of the performance metrics for the top 20 lab values, highlighting the consistent improvement of the Lab-MAE model over XGBoost. This visual analysis further supports the numerical results, emphasizing the Lab-MAE model's enhanced ability to predict missing lab values accurately.

\subsection{R2 metrics comparison per race of Lab-MAE for the top 20 lab values}

\begin{figure*}[ht!]
    \centering
    \includegraphics[width=1\linewidth]{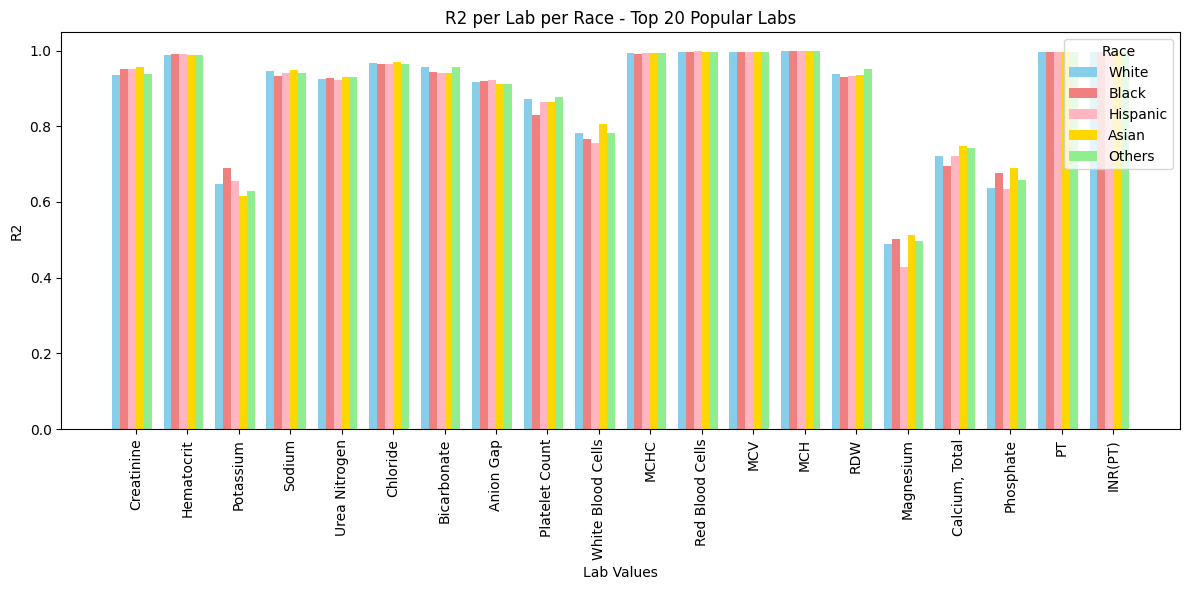}
    \caption{Comparison of the Lab-MAE model's performance across different races for the top 20 lab values. Blue indicates White race, red indicates black race, pink indicates Hispanic, yellow indicates Asian, and green indicates others.}
    \label{fig:top20_lab_values_per_race}
\end{figure*}

Figure \ref{fig:top20_lab_values_per_race} visually depicts the variation in \(R^2\) scores for the top 20 lab values across different racial groups. This visualization clearly demonstrates the disparities in model performance, with certain lab values consistently showing higher predictive accuracy for specific races while underperforming for others.

\clearpage
\section{Performance metrics across races}\label{apd:metrics-race}

\begin{table}[ht!]
    \centering
    \small
    \caption{Performance metrics (R\(^2\)) of Lab-MAE model across different racial groups for the top 20 lab values. The best-performing race for each lab value is highlighted in bold.}
    \begin{tabular}{lccccc}
        \toprule
        Lab ID & Asian & Black & Hispanic & Others & White \\
        \midrule
        Anion Gap & 0.9105 & 0.9197 & \textbf{0.9223} & 0.9110 & 0.9181 \\
        Bicarbonate & 0.9419 & 0.9431 & 0.9412 & \textbf{0.9577} & 0.9562 \\
        Calcium, Total & \textbf{0.7479} & 0.6950 & 0.7211 & 0.7439 & 0.7211 \\
        Chloride & \textbf{0.9702} & 0.9653 & 0.9647 & 0.9651 & 0.9671 \\
        Creatinine & \textbf{0.9565} & 0.9525 & 0.9522 & 0.9384 & 0.9354 \\
        Glucose & 0.4143 & 0.4669 & 0.4700 & \textbf{0.4823} & 0.4637 \\
        Magnesium & \textbf{0.5134} & 0.5028 & 0.4278 & 0.4965 & 0.4888 \\
        Phosphate & \textbf{0.6892} & 0.6773 & 0.6336 & 0.6581 & 0.6366 \\
        Potassium & 0.6151 & \textbf{0.6891} & 0.6566 & 0.6290 & 0.6487 \\
        Sodium & \textbf{0.9488} & 0.9327 & 0.9418 & 0.9397 & 0.9453 \\
        Urea Nitrogen & 0.9307 & 0.9276 & 0.9232 & \textbf{0.9313} & 0.9235 \\
        Hematocrit & 0.9884 & \textbf{0.9911} & 0.9910 & 0.9885 & 0.9889 \\
        Hemoglobin & 0.9983 & 0.9981 & \textbf{0.9985} & 0.9981 & 0.9983 \\
        MCH & \textbf{0.9982} & 0.9979 & 0.9975 & 0.9978 & 0.9977 \\
        MCHC & 0.9941 & 0.9922 & 0.9923 & 0.9936 & \textbf{0.9945} \\
        MCV & 0.9962 & \textbf{0.9966} & 0.9954 & 0.9958 & 0.9955 \\
        Platelet Count & 0.8654 & 0.8310 & 0.8644 & \textbf{0.8772} & 0.8712 \\
        RDW & 0.9365 & 0.9288 & 0.9334 & \textbf{0.9522} & 0.9381 \\
        Red Blood Cells & 0.9966 & 0.9964 & \textbf{0.9979} & 0.9968 & 0.9974 \\
        White Blood Cells & \textbf{0.8061} & 0.7672 & 0.7559 & 0.7822 & 0.7828 \\
        \bottomrule
    \end{tabular}
    \label{tab:top20_race_comparison}
\end{table}

\begin{table*}[ht]
    \centering
    \small
    \caption{Performance metrics (WD) of Lab-MAE model across different racial groups for the top 20 lab values. The best-performing race for each lab value is highlighted in bold.}
    \begin{tabular}{lccccc}
        \toprule
        Lab ID & Asian & Black & Hispanic & Others & White \\
        \midrule
        Anion Gap & 0.2159 & 0.2147 & \textbf{0.2011} & 0.2216 & 0.2058 \\
        Bicarbonate & 0.2126 & 0.2124 & 0.2071 & 0.2033 & \textbf{0.1944} \\
        Calcium, Total & 0.0872 & 0.0903 & 0.0890 & \textbf{0.0755} & 0.0818 \\
        Chloride & 0.2360 & 0.2387 & 0.2338 & 0.2330 & \textbf{0.2269} \\
        Creatinine & 0.0408 & 0.0471 & 0.0393 & 0.0360 & \textbf{0.0336} \\
        Glucose & 9.6687 & 11.1270 & 11.6503 & 10.5412 & \textbf{9.6597} \\
        Hematocrit & 0.0502 & 0.0423 & 0.0460 & 0.0455 & \textbf{0.0404} \\
        Hemoglobin & 0.0287 & 0.0272 & 0.0273 & 0.0267 & \textbf{0.0260} \\
        MCH & 0.0325 & 0.0297 & 0.0298 & 0.0307 & \textbf{0.0289} \\
        MCHC & 0.0291 & 0.0286 & 0.0283 & \textbf{0.0273} & 0.0279 \\
        MCV & 0.2389 & \textbf{0.2313} & 0.2411 & 0.2374 & 0.2402 \\
        Magnesium & 0.0731 & 0.0710 & 0.0692 & \textbf{0.0686} & 0.0718 \\
        Phosphate & \textbf{0.1378} & 0.1513 & 0.1588 & 0.1604 & 0.1572 \\
        Platelet Count & 8.8684 & 7.3407 & 7.2225 & \textbf{6.7014} & 7.7651 \\
        Potassium & \textbf{0.0926} & 0.0965 & 0.1005 & 0.0969 & 0.0954 \\
        RDW & 0.0829 & 0.0874 & 0.0920 & \textbf{0.0633} & 0.0703 \\
        Red Blood Cells & 0.0059 & 0.0050 & 0.0048 & 0.0046 & \textbf{0.0042} \\
        Sodium & 0.2176 & 0.2292 & 0.2276 & 0.2265 & \textbf{0.2167} \\
        Urea Nitrogen & 0.6366 & 0.9511 & 0.6031 & \textbf{0.5351} & 0.5394 \\
        White Blood Cells & 0.4670 & 0.5065 & 0.4525 & \textbf{0.4512} & 0.4537 \\
        \bottomrule
    \end{tabular}
    \label{tab:top20_race_comparison_wd}
\end{table*}

\begin{table*}[ht]
    \centering
    \small
    \caption{Performance metrics (RMSE) of Lab-MAE model across different racial groups for the top 20 lab values. The best-performing race for each lab value is highlighted in bold.}
    \begin{tabular}{lccccc}
        \toprule
        Lab ID & Asian & Black & Hispanic & Others & White \\
        \midrule
        Anion Gap & 0.8643 & 0.8356 & \textbf{0.7685} & 0.8601 & 0.7936 \\
        Bicarbonate & 0.8140 & 0.8678 & 0.8152 & 0.7607 & \textbf{0.7375} \\
        Calcium, Total & \textbf{0.3175} & 0.3435 & 0.3257 & 0.3192 & 0.3202 \\
        Chloride & 0.8745 & 0.9261 & 0.9079 & 0.9201 & \textbf{0.8565} \\
        Creatinine & 0.2467 & 0.2609 & 0.2293 & 0.2351 & \textbf{0.2239} \\
        Glucose & 30.9953 & 31.3567 & 32.5657 & \textbf{30.1562} & 28.1981 \\
        Hematocrit & 0.6032 & \textbf{0.5263} & 0.5282 & 0.5753 & 0.5719 \\
        Hemoglobin & 0.0805 & 0.0848 & \textbf{0.0750} & 0.0822 & 0.0779 \\
        MCH & \textbf{0.1042} & 0.1109 & 0.1137 & 0.1089 & 0.1080 \\
        MCHC & 0.1095 & 0.1267 & 0.1257 & 0.1148 & \textbf{0.1092} \\
        MCV & 0.3992 & \textbf{0.3792} & 0.4068 & 0.4038 & 0.4075 \\
        Magnesium & 0.1779 & 0.1783 & 0.1822 & \textbf{0.1763} & 0.1778 \\
        Phosphate & \textbf{0.5058} & 0.5182 & 0.5311 & 0.5104 & 0.5073 \\
        Platelet Count & 40.3788 & 41.9516 & 39.3368 & \textbf{38.1394} & 39.6089 \\
        Potassium & 0.2790 & 0.2724 & 0.2745 & 0.2779 & \textbf{0.2680} \\
        RDW & 0.5790 & 0.5828 & 0.5794 & \textbf{0.4920} & 0.5472 \\
        Red Blood Cells & 0.0401 & 0.0414 & \textbf{0.0317} & 0.0368 & 0.0335 \\
        Sodium & 0.8878 & 0.9470 & 0.8905 & 0.9367 & \textbf{0.8559} \\
        Urea Nitrogen & 4.7031 & 4.8157 & 4.5611 & \textbf{4.3308} & 4.4837 \\
        White Blood Cells & 1.9842 & 1.9612 & \textbf{1.9424} & 1.9685 & 1.9190 \\
        \bottomrule
    \end{tabular}
    \label{tab:top20_race_comparison_rmse}
\end{table*}


\clearpage
\section{Lab value distribution of Lab-MAE and XGBoost}\label{apd:4}
Prediction vs real distributions in test set [\href{https://drive.google.com/file/d/1dxiIT1mcKLU-GW3eUg5SSNpJ18V1jQsn/view?usp=sharing}{here}]

\section{Lab value distributions per race}\label{apd:5}

\begin{itemize}
    \item Prediction vs real distributions per race for Lab-MAE model in test set  [\href{https://drive.google.com/file/d/10jjOcPEC9lWC73pXRKDCMsCsGHHob4k_/view?usp=sharing}{here}]

    \item Prediction vs real distributions per race for XGBoost model in test set  [\href{https://drive.google.com/file/d/1ypoY2uxD98hGd3GM9Qz9ZKtcr28SjSVp/view?usp=drive_link}{here}]
\end{itemize}

\section{Lab value distributions with vs without follow-up values}\label{apd:6}

\begin{itemize}
    \item Prediction vs real distributions with vs without follow-up value in test set for Lab-MAE [\href{https://drive.google.com/file/d/1osyerCjf20QILMUKctpnR8t-X8n7CqMY/view?usp=sharing}{here}]

    \item Prediction vs real distributions with vs without follow-up value in test set for XGBoost [\href{https://drive.google.com/file/d/1XQ4m-qwA_mxAAZ-ks6Wl2zjvJ8RR90vU/view?usp=sharing}{here}]
\end{itemize}

\clearpage
\begin{thebibliography}{37}
\providecommand{\natexlab}[1]{#1}
\providecommand{\url}[1]{\texttt{#1}}
\expandafter\ifx\csname urlstyle\endcsname\relax
  \providecommand{\doi}[1]{doi: #1}\else
  \providecommand{\doi}{doi: \begingroup \urlstyle{rm}\Url}\fi

\bibitem[Austin et~al.(2021)Austin, White, Lee, and van Buuren]{austin2021missing}
Peter~C Austin, Ian~R White, Douglas~S Lee, and Stef van Buuren.
\newblock Missing data in clinical research: a tutorial on multiple imputation.
\newblock \emph{Canadian Journal of Cardiology}, 37\penalty0 (9):\penalty0 1322--1331, 2021.

\bibitem[Bao et~al.(2021)Bao, Dong, Piao, and Wei]{bao2021beit}
Hangbo Bao, Li~Dong, Songhao Piao, and Furu Wei.
\newblock Beit: Bert pre-training of image transformers.
\newblock \emph{arXiv preprint arXiv:2106.08254}, 2021.

\bibitem[Bellamy et~al.(2023)Bellamy, Kumar, Wang, and Beam]{bellamy2023labrador}
David~R Bellamy, Bhawesh Kumar, Cindy Wang, and Andrew Beam.
\newblock Labrador: Exploring the limits of masked language modeling for laboratory data.
\newblock \emph{arXiv preprint arXiv:2312.11502}, 2023.

\bibitem[Chen and Guestrin(2016)]{chen2016xgboost}
Tianqi Chen and Carlos Guestrin.
\newblock Xgboost: A scalable tree boosting system.
\newblock In \emph{Proceedings of the 22nd acm sigkdd international conference on knowledge discovery and data mining}, pages 785--794, 2016.

\bibitem[Courty et~al.(2024)Courty, Schmidt, Goyal-Kamal, MarionCoutarel, Blanche, Feld, inimaz, Lecourt, LiamConnell, SabAmine, supatomic, LLORET, Léval, Cruveiller, Saboni, ouminasara, Zhao, Joshi, Bauer, Bogroff, de~Lavoreille, Laskaris, Phiev, Abati, rosekelly6400, Blank, Wang, Otávio, and Catovic]{benoit_courty_2024_14518377}
Benoit Courty, Victor Schmidt, Goyal-Kamal, MarionCoutarel, Luis Blanche, Boris Feld, inimaz, Jérémy Lecourt, LiamConnell, SabAmine, supatomic, Patrick LLORET, Mathilde Léval, Alexis Cruveiller, Amine Saboni, ouminasara, Franklin Zhao, Aditya Joshi, Christian Bauer, Alexis Bogroff, Hugues de~Lavoreille, Niko Laskaris, Alexandre Phiev, Edoardo Abati, rosekelly6400, Douglas Blank, Ziyao Wang, Lucas Otávio, and Armin Catovic.
\newblock mlco2/codecarbon: v2.8.2, December 2024.
\newblock URL \url{https://doi.org/10.5281/zenodo.14518377}.

\bibitem[Devlin(2018)]{devlin2018bert}
Jacob Devlin.
\newblock Bert: Pre-training of deep bidirectional transformers for language understanding.
\newblock \emph{arXiv preprint arXiv:1810.04805}, 2018.

\bibitem[Dosovitskiy(2020)]{dosovitskiy2020image}
Alexey Dosovitskiy.
\newblock An image is worth 16x16 words: Transformers for image recognition at scale.
\newblock \emph{arXiv preprint arXiv:2010.11929}, 2020.

\bibitem[Du et~al.(2023)Du, Melis, and Wang]{du2023remasker}
Tianyu Du, Luca Melis, and Ting Wang.
\newblock Remasker: Imputing tabular data with masked autoencoding.
\newblock \emph{arXiv preprint arXiv:2309.13793}, 2023.

\bibitem[Grinsztajn et~al.(2022)Grinsztajn, Oyallon, and Varoquaux]{grinsztajn2022tree}
L{\'e}o Grinsztajn, Edouard Oyallon, and Ga{\"e}l Varoquaux.
\newblock Why do tree-based models still outperform deep learning on typical tabular data?
\newblock \emph{Advances in neural information processing systems}, 35:\penalty0 507--520, 2022.

\bibitem[Hastie et~al.(2015)Hastie, Mazumder, Lee, and Zadeh]{softimpute}
Trevor Hastie, Rahul Mazumder, Jason~D Lee, and Reza Zadeh.
\newblock Matrix completion and low-rank svd via fast alternating least squares.
\newblock \emph{The Journal of Machine Learning Research}, 16\penalty0 (1):\penalty0 3367--3402, 2015.

\bibitem[He et~al.(2022)He, Chen, Xie, Li, Doll{\'a}r, and Girshick]{he2022masked}
Kaiming He, Xinlei Chen, Saining Xie, Yanghao Li, Piotr Doll{\'a}r, and Ross Girshick.
\newblock Masked autoencoders are scalable vision learners.
\newblock In \emph{Proceedings of the IEEE/CVF conference on computer vision and pattern recognition}, pages 16000--16009, 2022.

\bibitem[Jarrett et~al.(2022)Jarrett, Cebere, Liu, Curth, and van~der Schaar]{hyperimpute}
Daniel Jarrett, Bogdan~C Cebere, Tennison Liu, Alicia Curth, and Mihaela van~der Schaar.
\newblock Hyperimpute: Generalized iterative imputation with automatic model selection.
\newblock In \emph{International Conference on Machine Learning}, pages 9916--9937. PMLR, 2022.

\bibitem[Johnson et~al.(2020)Johnson, Bulgarelli, Pollard, Horng, Celi, and Mark]{johnson2020mimic}
Alistair Johnson, Lucas Bulgarelli, Tom Pollard, Steven Horng, Leo~Anthony Celi, and Roger Mark.
\newblock Mimic-iv.
\newblock \emph{PhysioNet. Available online at: https://physionet. org/content/mimiciv/1.0/(accessed August 23, 2021)}, pages 49--55, 2020.

\bibitem[Johnson et~al.(2023)Johnson, Bulgarelli, Shen, Gayles, Shammout, Horng, Pollard, Hao, Moody, Gow, et~al.]{johnson2023mimic}
Alistair~EW Johnson, Lucas Bulgarelli, Lu~Shen, Alvin Gayles, Ayad Shammout, Steven Horng, Tom~J Pollard, Sicheng Hao, Benjamin Moody, Brian Gow, et~al.
\newblock Mimic-iv, a freely accessible electronic health record dataset.
\newblock \emph{Scientific data}, 10\penalty0 (1):\penalty0 1, 2023.

\bibitem[Kadra et~al.(2021)Kadra, Lindauer, Hutter, and Grabocka]{kadra2021well}
Arlind Kadra, Marius Lindauer, Frank Hutter, and Josif Grabocka.
\newblock Well-tuned simple nets excel on tabular datasets.
\newblock \emph{Advances in neural information processing systems}, 34:\penalty0 23928--23941, 2021.

\bibitem[Kingma(2013)]{kingma2013auto}
Diederik~P Kingma.
\newblock Auto-encoding variational bayes.
\newblock \emph{arXiv preprint arXiv:1312.6114}, 2013.

\bibitem[Kowsar et~al.(2024)Kowsar, Rabbani, and Samad]{kowsar2024attention}
Ibna Kowsar, Shourav~B Rabbani, and Manar~D Samad.
\newblock Attention-based imputation of missing values in electronic health records tabular data.
\newblock In \emph{2024 IEEE 12th International Conference on Healthcare Informatics (ICHI)}, pages 177--182. IEEE, 2024.

\bibitem[Krishnan et~al.(2022)Krishnan, Rajpurkar, and Topol]{krishnan2022self}
Rayan Krishnan, Pranav Rajpurkar, and Eric~J Topol.
\newblock Self-supervised learning in medicine and healthcare.
\newblock \emph{Nature Biomedical Engineering}, 6\penalty0 (12):\penalty0 1346--1352, 2022.

\bibitem[Lee and Kim(2023)]{lee2023self}
Do-Hoon Lee and Han-joon Kim.
\newblock A self-attention-based imputation technique for enhancing tabular data quality.
\newblock \emph{Data}, 8\penalty0 (6):\penalty0 102, 2023.

\bibitem[Li et~al.(2021)Li, Yan, Chaudhary, Avula, Mudiganti, Husby, Shahjouei, Afshar, Stewart, Yeasin, et~al.]{li2021imputation}
Jiang Li, Xiaowei~S Yan, Durgesh Chaudhary, Venkatesh Avula, Satish Mudiganti, Hannah Husby, Shima Shahjouei, Ardavan Afshar, Walter~F Stewart, Mohammed Yeasin, et~al.
\newblock Imputation of missing values for electronic health record laboratory data.
\newblock \emph{NPJ digital medicine}, 4\penalty0 (1):\penalty0 147, 2021.

\bibitem[Luo(2022)]{luo2022evaluating}
Yuan Luo.
\newblock Evaluating the state of the art in missing data imputation for clinical data.
\newblock \emph{Briefings in Bioinformatics}, 23\penalty0 (1):\penalty0 bbab489, 2022.

\bibitem[Luo et~al.(2016)Luo, Szolovits, Dighe, and Baron]{luo2016using}
Yuan Luo, Peter Szolovits, Anand~S Dighe, and Jason~M Baron.
\newblock Using machine learning to predict laboratory test results.
\newblock \emph{American journal of clinical pathology}, 145\penalty0 (6):\penalty0 778--788, 2016.

\bibitem[Panaretos and Zemel(2019)]{panaretos2019statistical}
Victor~M Panaretos and Yoav Zemel.
\newblock Statistical aspects of wasserstein distances.
\newblock \emph{Annual review of statistics and its application}, 6\penalty0 (1):\penalty0 405--431, 2019.

\bibitem[Parvaiz et~al.(2023)Parvaiz, Khalid, Zafar, Ameer, Ali, and Fraz]{parvaiz2023vision}
Arshi Parvaiz, Muhammad~Anwaar Khalid, Rukhsana Zafar, Huma Ameer, Muhammad Ali, and Muhammad~Moazam Fraz.
\newblock Vision transformers in medical computer vision—a contemplative retrospection.
\newblock \emph{Engineering Applications of Artificial Intelligence}, 122:\penalty0 106126, 2023.

\bibitem[Platias and Petasis(2020)]{softimpute_lab}
Christos Platias and Georgios Petasis.
\newblock A comparison of machine learning methods for data imputation.
\newblock In \emph{11th Hellenic Conference on Artificial Intelligence}, pages 150--159, 2020.

\bibitem[Renc et~al.(2024)Renc, Jia, Samir, Was, Li, Bates, and Sitek]{renc2024zero}
Pawel Renc, Yugang Jia, Anthony~E Samir, Jaroslaw Was, Quanzheng Li, David~W Bates, and Arkadiusz Sitek.
\newblock Zero shot health trajectory prediction using transformer.
\newblock \emph{NPJ Digital Medicine}, 7\penalty0 (1):\penalty0 256, 2024.

\bibitem[Riley et~al.(2024)Riley, Archer, Snell, Ensor, Dhiman, Martin, Bonnett, and Collins]{riley2024evaluation}
Richard~D Riley, Lucinda Archer, Kym~IE Snell, Joie Ensor, Paula Dhiman, Glen~P Martin, Laura~J Bonnett, and Gary~S Collins.
\newblock Evaluation of clinical prediction models (part 2): how to undertake an external validation study.
\newblock \emph{bmj}, 384, 2024.

\bibitem[Team(2024)]{codecarbon_global_energy_mix}
CodeCarbon Team.
\newblock Global energy mix data, 2024.
\newblock URL \url{https://github.com/mlco2/codecarbon/blob/master/codecarbon/data/private_infra/global_energy_mix.json}.
\newblock Accessed: December 25, 2024.

\bibitem[Teotia et~al.(2024)Teotia, Jia, Woite, Celi, Matos, and Struja]{teotia2024variation}
Khushboo Teotia, Yueran Jia, Naira~Link Woite, Leo~Anthony Celi, Jo{\~a}o Matos, and Tristan Struja.
\newblock Variation in monitoring: Glucose measurement in the icu as a case study to preempt spurious correlations.
\newblock \emph{Journal of Biomedical Informatics}, 153:\penalty0 104643, 2024.

\bibitem[van Breugel and van~der Schaar(2024)]{van2024tabular}
Boris van Breugel and Mihaela van~der Schaar.
\newblock Why tabular foundation models should be a research priority.
\newblock \emph{arXiv preprint arXiv:2405.01147}, 2024.

\bibitem[Vaswani(2017)]{vaswani2017attention}
A~Vaswani.
\newblock Attention is all you need.
\newblock \emph{Advances in Neural Information Processing Systems}, 2017.

\bibitem[Villani and Villani(2009)]{villani2009wasserstein}
C{\'e}dric Villani and C{\'e}dric Villani.
\newblock The wasserstein distances.
\newblock \emph{Optimal transport: old and new}, pages 93--111, 2009.

\bibitem[Waljee et~al.(2013)Waljee, Mukherjee, Singal, Zhang, Warren, Balis, Marrero, Zhu, and Higgins]{waljee2013comparison}
Akbar~K Waljee, Ashin Mukherjee, Amit~G Singal, Yiwei Zhang, Jeffrey Warren, Ulysses Balis, Jorge Marrero, Ji~Zhu, and Peter~DR Higgins.
\newblock Comparison of imputation methods for missing laboratory data in medicine.
\newblock \emph{BMJ open}, 3\penalty0 (8):\penalty0 e002847, 2013.

\bibitem[Wu et~al.(2020)Wu, Zhang, Ilyas, and Rekatsinas]{wu2020attention}
Richard Wu, Aoqian Zhang, Ihab Ilyas, and Theodoros Rekatsinas.
\newblock Attention-based learning for missing data imputation in holoclean.
\newblock \emph{Proceedings of Machine Learning and Systems}, 2:\penalty0 307--325, 2020.

\bibitem[Yoon et~al.(2018)Yoon, Jordon, and Schaar]{gain}
Jinsung Yoon, James Jordon, and Mihaela Schaar.
\newblock Gain: Missing data imputation using generative adversarial nets.
\newblock In \emph{International conference on machine learning}, pages 5689--5698. PMLR, 2018.

\bibitem[Zamanzadeh et~al.(2021)Zamanzadeh, Petousis, Davis, Nicholas, Norris, Tuttle, Bui, and Sarrafzadeh]{zamanzadeh2021autopopulus}
Davina~J Zamanzadeh, Panayiotis Petousis, Tyler~A Davis, Susanne~B Nicholas, Keith~C Norris, Katherine~R Tuttle, Alex~AT Bui, and Majid Sarrafzadeh.
\newblock Autopopulus: a novel framework for autoencoder imputation on large clinical datasets.
\newblock In \emph{2021 43rd Annual International Conference of the IEEE Engineering in Medicine \& Biology Society (EMBC)}, pages 2303--2309. IEEE, 2021.

\bibitem[Zhang et~al.(2020)Zhang, Yan, Gao, Malin, and Chen]{zhang2020predicting}
Xinmeng Zhang, Chao Yan, Cheng Gao, Bradley~A Malin, and You Chen.
\newblock Predicting missing values in medical data via xgboost regression.
\newblock \emph{Journal of healthcare informatics research}, 4:\penalty0 383--394, 2020.

\end{thebibliography}
\end{document}